 \documentclass[preprint,5p,times,twocolumn,authoryear]{elsarticle}

\usepackage{amssymb}
\usepackage{lipsum}
\usepackage{hyperref}
\hypersetup{
    colorlinks=true,
    linkcolor=blue,
    filecolor=magenta,      
    urlcolor=cyan,
    pdftitle={Overleaf Example},
    pdfpagemode=FullScreen,
    }
\usepackage{subcaption}
\usepackage{enumitem}
\usepackage{graphicx}
\begin{document}

\begin{frontmatter}



\title{FairyLandAI: Personalized Fairy Tales utilizing ChatGPT and DALLE-3}


\author[first]{Georgios Makridis}
\author[second]{Athanasios Oikonomou}
\author[first]{Vasileios Koukos}

\affiliation[first]{organization={University of Piraeus, Department of Digital Systems},
           addressline={Karaoli ke Dimitriou 80}, 
           city={Piraeus},
           postcode={18534}, 
           state={Attica},
           country={Greece}}
\affiliation[second]{organization={National and Kapodistrian University of Athens},
           addressline={}, 
           city={Athens},
           postcode={}, 
           state={Attica},
           country={Greece}}

\begin{abstract}
In the diverse world of AI-driven storytelling, there is a unique opportunity to engage young audiences with customized, and personalized narratives. This paper introduces FairyLandAI an innovative Large Language Model (LLM) developed through OpenAI's API, specifically crafted to create personalized fairytales for children. The distinctive feature of FairyLandAI is its dual capability: it not only generates stories that are engaging, age-appropriate, and reflective of various traditions but also autonomously produces imaginative prompts suitable for advanced image generation tools like GenAI and Dalle-3, thereby enriching the storytelling experience. FairyLandAI is expertly tailored to resonate with the imaginative worlds of children, providing narratives that are both educational and entertaining and in alignment with the moral values inherent in different ages. Its unique strength lies in customizing stories to match individual children's preferences and cultural backgrounds, heralding a new era in personalized storytelling. Further, its integration with image generation technology offers a comprehensive narrative experience that stimulates both verbal and visual creativity. Empirical evaluations of FairyLandAI demonstrate its effectiveness in crafting captivating stories for children, which not only entertain but also embody the values and teachings of diverse traditions. This model serves as an invaluable tool for parents and educators, supporting them in imparting meaningful moral lessons through engaging narratives. FairyLandAI represents a pioneering step in using LLMs, particularly through OpenAI's API, for educational and cultural enrichment, making complex moral narratives accessible and enjoyable for young, imaginative minds.
\end{abstract}




\end{frontmatter}

%\tableofcontents



\section{Introduction}
\label{introduction}

In the Information Age, characterized by the proliferation of advanced computational systems and significant data generation, we witness the evolution of the fourth industrial revolution, Industry 4.0  \cite{makridis2020predictive}. At its heart lies Artificial Intelligence (AI), propelling the development of innovative tools and methods. Concurrently, there is a surge in interest towards Explainable AI (XAI), aiming to demystify machine learning decisions.

This paper introduces FairyLandAI\footnote{Visit  \href{https://fairylandai.com}{fairylandai.com} or download our \href{https://play.google.com/store/apps/details?id=com.fairylandai}{Android application}.}, a novel GPT-based Large Language Model (LLM) utilizing OpenAI's API, specifically designed for crafting personalized fairytales for children. Unlike previous models, FairyLandAI focuses on child-centric narrative generation and integrates with Dalle-3 for the creation of visually consistent characters, contributing significantly to the field of AI in storytelling and education. The salient features of FairyLandAI include:

\begin{enumerate}
    \item \textbf{Child-Centric Narrative Creation}: Leveraging Large Language Models, FairyLandAI generates engaging and appropriate stories that align with the diverse moral landscapes of various traditions, tailored to the unique preferences of each child.
    
    \item \textbf{Integration with Dalle-3 for Character Consistency}: A key innovation is the model's ability to generate prompts for Dalle-3, ensuring visual consistency in character design across stories. This feature enhances the visual storytelling experience, maintaining continuity and depth in character portrayal.
    
    \item \textbf{Educational and Moral Framework}: The app embeds educational and moral dimensions within narratives, aiding in the cognitive and ethical development of children, a critical aspect for parents and educators.

    \item \textbf{Polyglot - Multilingual}: Based on FairylandAI developed on top of openai chatgpt, we have leveraged many capabilities of it. One of them is the capability of being Multilingual.  
    
    \item \textbf{Empirical Validation and Feedback}: The efficacy of FairyLandAI is validated through empirical studies, including user feedback from children, parents, and educators, emphasizing its impact in delivering engaging, educational, and ethically rich stories.
\end{enumerate}

FairyLandAI represents an innovative approach in AI-driven storytelling, merging technology with child-centric educational objectives, and advancing the field of personalized and moral storytelling for young audiences.

The remainder of the paper is organized as follows: Section 2 presents the background and the motivation of our research, while Section 3 delivers the literature review in the areas of study of this paper. Section 4, presents the proposed methodological approach, introduces the overall implementation, and offers details regarding the datasets used and the evaluation procedure. Section 6 dives deeper into the results of the conducted research and the corresponding survey. Section 5 concludes with recommendations for future research and the potential of the current study.

\section{Background}
\par Our research's underlying motivation is illuminated through an introduction to the foundational concepts of Image Classification, XAI methods, and adversarial attacks.

\subsection{Traditional Storytelling}
With the evolution of interactive storytelling as a key medium in the Digital Age, there is an emerging need for advanced computational models capable of crafting personalized and interactive narratives. Interactive storytelling holds immense potential for unique, user-tailored experiences. Previous research in modeling character attributes and goals, as well as character points of view \cite{riedl2009incorporating}, has expanded opportunities for personalized narratives. However, these models often face limitations due to their predefined range of character traits and story arcs, restricting the depth of personalization possible in interactive storytelling.

"FairyLandAI" seeks to overcome these limitations by offering a more dynamic and flexible approach to storytelling. This model diverges from traditional systems by allowing extensive customization of character traits and narrative elements. These attributes can be dynamically adjusted to significantly influence the story's progression. This capability is particularly crucial in creating fairy tales that resonate with children, where personalization includes not only the plot but also the integration of diverse character and setting elements, reflecting the imaginative needs of young audiences.

Additionally, "FairyLandAI" enhances storytelling by integrating with Dalle-3 for the generation of consistent and contextually appropriate image prompts. This feature adds a visual layer to the storytelling experience, ensuring that the visual elements align seamlessly with the evolving narrative and character profiles. By addressing these essential aspects, "FairyLandAI" represents a significant advancement in the field of interactive storytelling, providing a more immersive and personally relevant experience for children, aligning with the advancements discussed in recent scholarly works.

\subsection{From Traditional to Digital Fairy Tales}

According to \cite{stavroupersonalized}, fairy tales hold significant educational and therapeutic potential, especially for children with autism in special education. These narratives enable children to identify with the hero, providing a framework for understanding and resolving personal conflicts. The symbolic language of fairy tales reaches deep into the unconscious, facilitating emotional and cognitive development. They serve not only as a source of entertainment but also as a means of personal growth, teaching lessons in morality, psychology, and mental resilience. While our current research has not yet delved into the full implications of fairy tales in special education, particularly for autistic children, it presents a promising area for future exploration. The unique blend of storytelling and symbolic interpretation in fairy tales could offer valuable insights into developmental approaches and therapeutic techniques in this field.

The advent of AI-generated personalized narratives, as exemplified by "FairyLandAI," brings forth a pivotal challenge: understanding how children form their sense of self within these AI-personalized texts. This paper takes a conceptual approach to explore this emerging domain, particularly in light of the profound shift in children's interactions with media and narratives. Drawing on Simon Garfield's metaphor of spatial understanding (Garfield, 2013: 19) and the insights of Wakenshaw and Dhamotharan (2019) on digital personalization.

This shift, indicative of a broader change in children’s learning and development, is characterized by narratives that are increasingly centered around personal data and preferences. The move from generalized media representations to highly individualized experiences, as noted by Van Kleeck et al. (1997), suggests significant implications for cognitive and identity development in children. Our research aims to shed light on how these AI-personalized narratives impact the way children engage with and make sense of their world, marking a crucial area in the evolution of digital literacy and child development.

\subsection{Challenges in AI-based digital child-book generation}

\cite{kucirkova2020digital}
A key challenge in the realm of AI-generated personalized fairy tales, as exemplified by "FairyLandAI," lies in balancing the enhancement of digital literacies with the impact on children’s developing sense of self. This issue becomes particularly pertinent when assessing the potential implications of integrating artificial intelligence (AI) in storytelling. "FairyLandAI" diverges from traditional personalized books by crafting unique stories based on comprehensive personalization parameters, potentially extending to AI-generated images that align with the narrative.

The challenge centers on navigating the interplay between the objective space (A-to-B), where AI crafts a story journey with unknown elements, and the subjective space (me-to-B), where the narrative is intimately tailored to the child’s individual world. This balance is crucial in maintaining a healthy development of self-identity in young readers. The personalization offered by "FairyLandAI" potentially blurs the boundaries between these objective and subjective experiences, leading to novel questions about the formation and understanding of self in children.

Therefore, a critical aspect of our research is to investigate how AI-driven personalization in storytelling influences children’s perception of their own identities. While AI offers unprecedented levels of story customization, it also poses challenges in how children interact with and internalize these narratives. The goal is to ensure that AI-enhanced personalized stories foster positive identity development, avoiding scenarios where the technology might inadvertently diminish the child’s agency in shaping their own sense of self.

\subsection{Objectives}
%% Objectives content

This paper sets out to redefine the landscape of personalized storytelling for children through the development and implementation of ``FairyLandAI,'' a novel framework that harnesses the power of Large Language Models (LLMs) and DALLE-3 image generation. Our objectives are multifaceted, aiming to blend technological innovation with educational and moral enrichment, thereby creating a new paradigm in children's literature that is both engaging and instructive. The core aims of our research are as follows:

\begin{enumerate}
    \item \textbf{To Develop a Child-Centric Narrative Generation System}: At the heart of FairyLandAI is the ambition to create stories that are not only captivating but also tailored to the individual preferences, cultural backgrounds, and ethical values of each child. This system leverages the capabilities of LLMs to produce narratives that are diverse, inclusive, and reflective of the moral landscapes of various traditions.
    
    \item \textbf{To Integrate Advanced Image Generation for Enhanced Storytelling}: Recognizing the importance of visual elements in children's stories, FairyLandAI incorporates DALLE-3 to generate consistent and contextually appropriate images that complement the narrative. This integration aims to enrich the storytelling experience, making it more immersive and visually stimulating for young audiences.
    
    \item \textbf{To Embed Educational and Moral Frameworks within Narratives}: Beyond entertainment, FairyLandAI seeks to imbue stories with educational content and moral lessons that align with the values and teachings of different cultures and traditions. This objective underscores our commitment to leveraging technology for cognitive and ethical development, offering a novel tool for parents and educators.
    
    \item \textbf{To Conduct Empirical Validation and Gather Feedback}: A crucial aspect of our research involves rigorous empirical evaluations of FairyLandAI, incorporating feedback from children, parents, and educators to assess the effectiveness, engagement, and educational impact of the generated stories. This iterative process is designed to refine and enhance the model based on user experiences and insights.
    
    \item \textbf{To Explore Multilingual Capabilities}: Acknowledging the global diversity of children's cultural and linguistic backgrounds, FairyLandAI aims to be polyglot, offering personalized storytelling across multiple languages. This expands the reach and inclusivity of the project, making it accessible to a wider audience.
\end{enumerate}

Based on these objectives, our research is driven by a holistic approach that combines cutting-edge AI technologies with deep insights into child psychology, education, and storytelling traditions. Through FairyLandAI, we aim to pioneer a new frontier in digital education and personalized entertainment, where technology serves as a bridge between traditional narratives and modern, inclusive storytelling that resonates with today's diverse, global audience.

\section{Literature Review}
%% Your Literature Review content

Recent advancements in AI-generated content (AIGC) have significantly enhanced the capabilities of digital storytelling, offering new avenues for creating personalized and engaging narratives. Innovative approaches in AIGC, as explored in "Innovative Digital Storytelling with AIGC: Exploration and Discussion of Recent Advances," emphasize the critical role of human intervention in refining AI-generated outputs to ensure high-quality, consistent character portrayal and visual coherence. This research highlights the use of advanced tools such as Midjourney and Runway for image generation and post-processing, ensuring that the visual elements align with the narrative's context and aesthetic preferences \cite{gu2023innovative}.

Furthermore, the study "Generative AI-Driven Storytelling: A New Era for Marketing" illustrates how generative AI can craft personalized narratives that resonate deeply with audiences, suggesting potential applications in real-time storytelling and immersive experiences, which align with the objectives of FairyLandAI \cite{vidrih2023generative}. Including these insights underscores the transformative potential of AI in creating bespoke storytelling experiences, bridging technical innovation with creative expression. This integration of AIGC into digital storytelling not only enhances the narrative and visual quality but also introduces novel methods for audience engagement and content personalization \cite{gu2023innovative}, \cite{vidrih2023generative}.

\subsection{Digital Education: A Technical Overview}
%%\label{}
The advent of digital technologies has ushered in a new era for educational paradigms, significantly enriching the learning experience with innovative tools and methodologies. The essence of digital education lies in leveraging computational systems to facilitate and enhance learning processes, encompassing e-learning platforms, Massive Open Online Courses (MOOCs), digital content, and interactive learning environments (\cite{bates2019teaching}; \cite{dziubaniuk2023learning}). These technologies have not only expanded access to education but also introduced novel pedagogical strategies that personalize the learning experience.

Artificial Intelligence (AI) and Machine Learning (ML) stand at the forefront of this transformation, offering sophisticated mechanisms for adaptive learning and personalized education. AI-driven systems analyze learners' behavior, preferences, and performance to tailor educational content, thereby addressing individual learning needs in many contexts, for example in training the workers in manufacturing environments \cite{peres2020industrial}. This personalization extends to intelligent tutoring systems, which simulate one-on-one tutoring by providing feedback, scaffolding learning tasks, and adjusting instructional strategies based on the learner's progress \cite{nkambou2023learning}.

Moreover, the integration of Natural Language Processing (NLP) technologies has facilitated the development of virtual assistants and chatbots that support learners by answering queries, offering explanations, and guiding them through learning materials, making education more interactive and accessible (\cite{hien2018intelligent}). These AI-enabled tools exemplify the potential of digital education to transcend traditional educational boundaries, offering scalable and effective learning solutions.

Cloud computing also plays a pivotal role in digital education, enabling the delivery of content and services through the Internet. This technology supports the scalability and accessibility of learning resources, ensuring that students and educators can access educational materials anytime and anywhere, fostering a flexible learning environment \cite{qasem2019cloud}.

Despite these advancements, the digital divide remains a significant challenge, highlighting disparities in access to digital technologies based on socioeconomic status, geography, and infrastructure \cite{lythreatis2022digital}. Furthermore, concerns around data privacy, ethical considerations in AI, and the need for digital literacy skills among both educators and learners are critical issues that must be addressed to realize the full potential of digital education \cite{akgun2022artificial}.

In summary, digital education technologies, powered by AI and cloud computing, offer promising avenues for enhancing and personalizing learning experiences. However, for these technologies to be effectively and equitably integrated into educational systems, it is essential to navigate the challenges associated with access, ethics, and implementation \cite{weller202025}.

\subsection{Personalized Books for Children}
Personalized books are books uniquely designed for individual children. The extent of personalization ranges from books created or selected by a digital system to match a child’s age or expressed reading interest, to sophisticated individualized books that use personal attributes (such as children’s names, drawings, or photographs) to tailor story characteristics, aesthetics and/or the story plot. Most paper-based personalized books give agency to the parent in that they ask them to supply the child’s data and select
the story to be particularized on a digital platform and then printed on demand (for example, Wonderbly Ltd). Digital storybooks have broadened the options, as young readers can personalize story illustrations, as well as audio and interactive features. For example, with the Mr. Glue digital personalized stories, children can add their names, audio recordings, and digital drawings to the story. Some digital personalized books give agency to the child and invite children to create their own stories based on their content (for example, the Our Story app), and some provide children with story templates to complete according to the publisher’s expectations (for example, the Nosy Crow Fairytale app). All these examples concern static personalization, that is, the use of personal data that does not change during the reading experience. Some digital personalized books, however, can be enhanced with data-collecting features that record children’s engagement, history and progress and, based on these data, recommend new content. Such dynamic personalization is run by algorithms and artificial intelligence (AI) that adapt the reading experience to an individual reader as
the book is read, simultaneously and seamlessly during the reading process, in ways comparable to how the blue dot aligns the focus of the digital map as its user moves around.

\subsection{Language Models in AI}

Since the advent of GPT, a variety of Large Language Models (LLMs) have been developed, showcasing remarkable skills in a range of Natural Language Processing (NLP) tasks, particularly in the financial sector.

A prominent example in this area is BloombergGPT, devised by Bloomberg's AI team, which has been trained with a vast array of financial documents. This model has demonstrated superior performance in financial NLP tasks \citep{wu2023bloomberggpt}. As of May 2023, BloombergGPT is primarily used internally at Bloomberg and does not have a public API.

Another significant LLM in this field is Google's Bard, a direct rival to ChatGPT. Utilizing Google's LAMDA (Language Model for Dialogue Applications), Bard integrates elements of BERT and GPT for creating dynamic, context-sensitive dialogues \citep{thoppilan2022lamda}. Similar to BloombergGPT, Bard has not released an open API as of the time of this writing.

Additionally, BLOOM, a competitor to GPT-3 in the open-source arena \citep{scao2022bloom}, has caught the attention of the LLM community. Although it's open-source, effectively deploying BLOOM demands substantial technical expertise and computational resources, and it does not have a version specifically tailored for conversational tasks, an area where models like ChatGPT shine.

Post the introduction of ChatGPT, a myriad of LLMs have surfaced, each focusing on specific functionalities like code completion \citep{dakhel2023github}, content creation, and marketing. These models provide specialized capabilities, broadening the applications and influence of LLMs. ChatGPT remains a leader in this domain \citep{jasper23report}, attributed to its public API, comprehensive training data, and its adaptability in diverse tasks. While ChatGPT is widely used in sectors such as healthcare and education \citep{sallam2023chatgpt}, its direct application in financial sentiment analysis is relatively less explored. \cite{fatouros2023transforming} indicates that ChatGPT can grasp complex contexts that require advanced reasoning skills, even with zero-shot prompting. Furthermore, MarketSense-AI, a practical financial tool, employs GPT-4 with the Chain-of-Thought (CoT) approach for elucidating investment decisions effectively \cite{fatouros2024can}. [

Finally another interesting utility of LLMs and specifically GPT-4 model, is the one presented by \cite{mavrepis2024xai} where it is proposed a Human-centric XAI tool. The paper introduces "x-[plAIn]", a new approach in Explainable Artificial Intelligence (XAI), developed to make XAI more accessible to non-experts. It utilizes a custom LLM to create clear, audience-specific summaries of XAI methods for different groups like business professionals and academics. The model adapts explanations to the audience's knowledge level and interests, improving decision-making and accessibility. Use-case studies show its effectiveness in making complex AI concepts understandable to a diverse range of users.

\subsection{Large Language Models in Digital Education}
The integration of Large Language Models (LLMs) into digital education systems has marked a revolutionary shift in the way educational content is delivered, personalized, and interacted with. These models, exemplified by OpenAI's GPT series, have demonstrated unparalleled capabilities in understanding and generating human-like text, opening new avenues for educational applications \cite{brown2020language}.

Education's critical role in societal progress is juxtaposed with traditional challenges, including diverse student needs and resource constraints, spotlighting Large Language Models' (LLMs) potential in revolutionizing digital learning. This evolving field of educational LLMs (EduLLMs) offers novel solutions for personalized instruction and assessment, aiming to enhance educational quality and experiences, with ongoing research providing essential insights and directions for leveraging LLMs in smart education.\cite{gan2023large}

Also as described in \cite{generationrole} several gaps where pinpointed in the utilization of LLMs for creating personalized educational tools. While models like GPT-4 demonstrate capabilities in generating accurate questions, there is a notable underutilization of Retrieval-Augmented Generation (RAG) models and self-operating tools, which could significantly enhance the personalization of learning experiences.

These studies collectively illustrate the evolving role of LLMs in the generation of educational content, offering insights into their capabilities to support educators and improve learning outcomes. The continuous advancements in LLM technologies promise further enhancements in personalized education and assessment strategies.

Recent studies have underscored the effectiveness of Large Language Models (LLMs) like GPT-4, GPT-3, BERT, and GPT-2 in enhancing the generation of educational content, including multiple-choice and short-answer questions. These advancements demonstrate the potential of LLMs to significantly contribute to personalized learning experiences and assessment methodologies.

\begin{itemize}
    \item \textbf{Generating Multiple Choice Questions for Computing Courses using Large Language Models} \cite{tran2023generating}: This study highlights that GPT-4 outperforms GPT-3 in generating accurate multiple-choice questions, marking a significant improvement in the quality of AI-generated educational assessments.
    
    \item \textbf{Short Answer Questions Generation by Fine-Tuning BERT and GPT-2 } \cite{tsai2021short}: The fine-tuning of AI models like BERT and GPT-2 has been shown to assist educators in automatically generating short-answer exam questions, facilitating a more efficient assessment creation process.
    
    \item \textbf{Assessing the Quality of Student-Generated Short Answer Questions Using GPT-3} \cite{moore2022assessing}: An analysis reveals that 32\% of student-generated short answer questions are of high quality. This study explores the use of GPT-3 for evaluating and potentially improving the caliber of student-created questions.
    
    \item \textbf{Reading Comprehension Quiz Generation using Generative Pre-trained Transformers }\cite{dijkstra2022reading}: EduQuiz employs AI technologies like GPT-3 to create quizzes that enhance learning practices and student engagement, showcasing the application of LLMs in creating dynamic and interactive educational content.
    
    \item \textbf{Leveraging GPT-3 as a question generator in Swedish for High School teachers} \cite{goran2023leveraging}: This research assesses the utility of GPT-3 in question creation for Swedish High School teachers through expert interviews, indicating the model's versatility and effectiveness across different languages and educational contexts.
\end{itemize}

\section{Methodology}

\begin{figure*}[h!]
\centering
\includegraphics[width=0.8\textwidth]{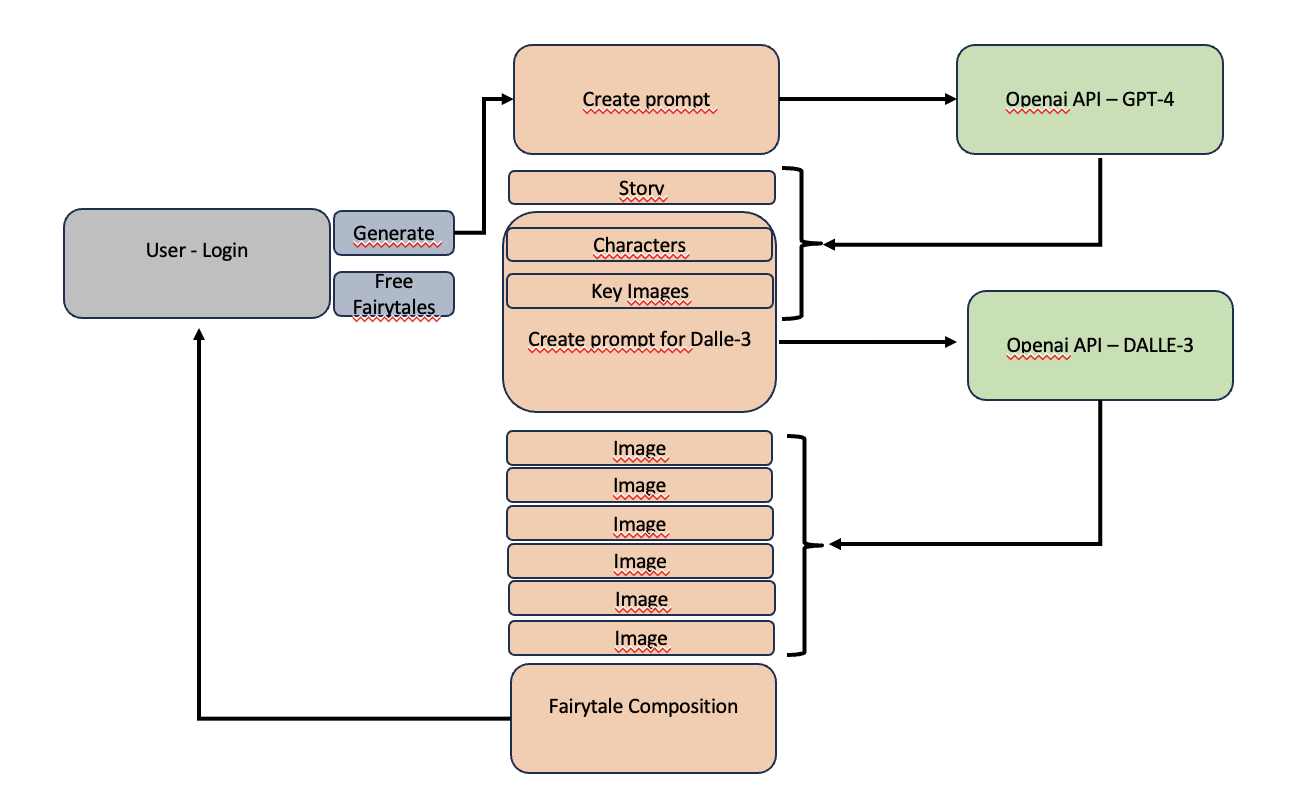}
\caption{ Process Diagram Generating a custom fairytale.}
\label{fig:trajectory}
\end{figure*}

FairyLandAI's architectural framework, illustrated in Figure 1, seamlessly integrates multiple core components, each meticulously developed using OpenAI's API and employing the GPT-4 model (OpenAI, 2023). This model is adept at using one-shot prompting and in-context learning to perform distinct tasks, essential for the creation of personalized fairytales for children and generating prompts for character creation in Dalle-3 \cite{betker2023improving}.

The framework is designed to mirror the cognitive and creative processes typical of storytelling and character development in children's literature. It encompasses various components including a narrative generator for tailoring fairytale stories, a character concept generator for creating consistent character images using Dalle-3, and a cultural and ethical values integrator to ensure that the stories align with diverse traditions and moral teachings.

Specifically, the narrative generator component focuses on crafting engaging and age-appropriate stories, leveraging advanced natural language processing techniques. The character concept generator, integrated with Dalle-3, ensures visual consistency and richness in character portrayal across different stories, enhancing the visual appeal and engagement for young readers. Additionally, the cultural and ethical values integrator ensures that the narratives align with the moral landscapes of various traditions, making the stories not only entertaining but also educational.

This holistic approach, combining AI-driven narrative generation with advanced image generation, sets FairyLandAI apart in the realm of AI-driven storytelling. It emulates the intricate process of crafting compelling and morally enriching stories for children, akin to the methods used by skilled storytellers and illustrators

\subsection{GPT-based Storytelling}

In the development of our GPT-based model for "FairyLandAI", we adopted a rigorous and technical approach tailored to the unique demands of crafting personalized fairytales and generating consistent character imagery for children. The initial phase involved defining the model's core objectives through a specialized interface, with a focus on generating narrative content that is both engaging and age-appropriate, and creating prompts for character consistency in Dalle-3 image generation. Our configuration phase was extensive, involving meticulous customization of the model’s parameters. This included naming, operational descriptions, and the development of initialization prompts tailored for generating child-centric narratives and corresponding image prompts, as outlined in Table \ref{tab:refined_prompt_evolution}.

Adherence to prompt-engineering guidelines, as recommended in the official OpenAI documentation, was a crucial aspect of our development process, accessible via this \href{https://platform.openai.com/docs/guides/prompt-engineering}{hyperlink}. Recognizing the importance of creating stories and images that are coherent and culturally sensitive, we incorporated an approach that balances creativity with ethical considerations. Instead of conventional methods, our strategy involved integrating narrative elements that foster moral and educational values, a process validated in studies such as \cite{tan2023educational}. This methodological rigor ensured the development of a GPT model finely tuned for storytelling and image prompt generation, enhancing its capability to deliver narratives and visualizations that are both captivating and meaningful for young audiences.

\subsection{GPT-based prompts for Dalle-3}

The creation of GPT-based prompts for Dalle-3 imagery within "FairyLandAI" necessitated a focused and iterative approach, designed to meet the dual goals of producing visually compelling and contextually accurate images to accompany the generated fairy tales. This process was centered on fine-tuning the GPT model to generate prompts that Dalle-3 could interpret effectively, ensuring that the resulting images would be both relevant to the story and resonate with a young audience. 

Our approach was grounded in a deep analysis of narrative elements that could translate effectively into visual representations. This involved the development of a detailed framework for prompt engineering, within the prompt of storytelling, aimed at optimizing the compatibility between textual prompts and the Dalle-3 image generation process, as summarized in Table \ref{tab:refined_prompt_evolution}.

Adherence to best practices in prompt engineering, as outlined in our methodological framework, was pivotal. We leveraged insights from leading research and the OpenAI documentation to refine our prompt generation process, available via this \href{https://platform.openai.com/docs/guides/prompt-engineering}{hyperlink}. Our methodology was geared towards ensuring that each prompt would not only generate aesthetically pleasing images but also maintain a strong alignment with the story's narrative and ethical themes, thereby enhancing the overall storytelling experience for children.

\begin{table*}[htbp]
\scriptsize
\centering
\caption{Refined Evolution of Fairy Tale Generation Prompts with Specific Changes}
\label{tab:refined_prompt_evolution}
\begin{tabular}{|l|p{11cm}|p{4cm}|}
\hline
\textbf{Version} & \textbf{Prompt As-Is} & \textbf{Specific Change from Previous Version} \\
\hline
V0 & ChatGPT, from now on you are going to act as an experienced fairy tale writer for kids. You will write a novel and fascinating fairy tale of more than 1000 words length. The fairy tale should be comprehensive and friendly for a 5 years old boy and the theme will be for cars... & Initial comprehensive prompt setting the task for creating a fairy tale with a focus on cars for a 5-year-old audience. \\
\hline
P1 & Craft a fairy tale story suitable for children. Include characters, setting, and moral of the story. Ensure the story is engaging and age-appropriate. Additionally, generate visual prompts for key scenes. Return the output in JSON format. & Establishes the foundational structure for story generation, including visual prompts for key scenes, and introduces the JSON output format. \\
\hline
P2 & Craft a fairy tale story suitable for children. Include characters, setting, and moral of the story. Ensure the story is engaging and age-appropriate. Additionally, generate visual prompts for key scenes to align with the narrative. Refine character and setting descriptions to accurately guide the generation of image prompts. Emphasize consistency in age, appearance, and cultural background. Return the output in JSON format with variables such as age, gender, element\_1, and element\_2. & Improves narrative detail and visual consistency, emphasizing accurate and contextually appropriate imagery, and introduces variable placeholders. \\
\hline
P3 & Craft a fairy tale story suitable for children. Include characters, setting, and moral of the story. Ensure the story is engaging and age-appropriate. Additionally, generate visual prompts for key scenes to align with the narrative. Refine character and setting descriptions to accurately guide the generation of image prompts. Emphasize consistency in age, appearance, and cultural background. Incorporate user feedback to customize character traits, setting details, and moral themes based on preferences. Return the output in JSON format with variables such as age, gender, element\_1, and element\_2. & Adds user customization to enhance engagement and relevance, tailoring stories to specific preferences, and maintains JSON output with variables. \\
\hline
P4 & Craft a fairy tale story suitable for children. Include characters, setting, and moral of the story. Ensure the story is engaging and age-appropriate. Additionally, generate visual prompts for key scenes to align with the narrative. Refine character and setting descriptions to accurately guide the generation of image prompts. Emphasize consistency in age, appearance, and cultural background. Incorporate user feedback to customize character traits, setting details, and moral themes based on preferences. Ensure all image prompts are consistent in style and character appearance. Provide descriptions for four key scenes, specifying details for DALLE-3. Return the output in JSON format with variables such as age, gender, element\_1, and element\_2. & Enhances image consistency and provides detailed scene descriptions for DALLE-3, ensuring visual coherence, and maintains JSON output with variables. \\
\hline
P5 & You are an experienced fairytale author for children designed to output JSON. Craft an engaging fairytale that is approximately words\_length words in length. The story should be geared towards a age-year-old gender, with a particular focus on element\_1. Use language comprehensively for a age-year-old child. The plot should contain element\_2 elements that stimulate the reader's imagination and conclude on a happy note. Provide descriptions, in English, for four key scenes from your story that will be used as prompts for DALLE-3 to generate the illustrations. These image descriptions should be more than fifty words in length each, and MUST be suitable for a age-year-old child. The prompts that you will provide should ensure that ALL the images are consistent in terms of style, and characters' appearance. The setting or background of each scene should be detailed as well. You should use simple words that can be pictured clearly without doubt, describe visually well-defined objects, speak in positives, specify what you want clearly, try to use singular nouns or specific numbers, and avoid concepts that involve significant extrapolation. For every image description, please follow the following format without using capital letters except for the characters' names: style render of [ONLY the name of the main characters and doing and positioning] and [the location and the background details], [description of non-main characters], image\_specs. Also, describe an image for the book cover that will be indicative of the fairytale, and MUST be suitable for a age-year-old child. Also give the FULL, detailed and extensive description of ALL the characters that are named in the story (specifically their detailed appearance, age, gender, form, style, human or animal, etc ) that are referenced by name in the story, in English. The description of the appearance of each character should be around 30 words in length. The style and form should be clear and not ambiguous to be visualized. If the title contains names, be sure to use names from the characters mentioned in the story. Title: the title, Characters: [{all characters information}], Story: the story, Image Descriptions: {ImageX: description}, Book Cover Description: the description of the cover. Do not use and use UTF-8 encoding. Please ignore the following timestamp timestamp. & Final comprehensive prompt integrating all aspects of storytelling, image generation, consistency, and detailed descriptions. \\
\hline
\end{tabular}
\end{table*}

\subsection{Audience Analysis and Content Customization}
Audience analysis plays a pivotal role in tailoring the storytelling experience provided by "FairyLandAI". By understanding the demographic, cultural background, and interests of the target audience, the model is adept at customizing content to meet the specific needs and preferences of individual children. This customization extends to the selection of themes, moral lessons, and even the complexity of the language used, ensuring that the stories are not only engaging but also accessible to the intended audience. Content customization, therefore, stands as a cornerstone of "FairyLandAI", enabling it to deliver personalized and meaningful narrative experiences that resonate with diverse audiences.

\subsection{Prompt engineering}

The iterative refinement of prompts presented in this study underscores the nuanced approach required to harness AI's potential in creating child-friendly fairy tales and illustrations. As demonstrated through the prompt evolution from V0 to P5, significant emphasis was placed on enhancing narrative coherence, thematic relevance, and visual consistency to engage a young audience effectively.

One of the most critical observations from this progression is the increasing specificity and clarity of the prompts. This specificity not only aids the AI in generating content that is more aligned with the intended thematic and educational goals but also ensures that the resultant tales and images are tailored to the cognitive and emotional level of the target age group. By introducing variables and detailed instructions, we cater to a broad spectrum of themes and settings, thereby expanding the diversity and inclusivity of the content generated.

Moreover, the introduction of steps in P3 and further refinement in P4 and P5 highlight a significant shift toward a more structured and phased approach to content creation. This method facilitates a deeper engagement with the narrative and illustrative elements, allowing for a more iterative and feedback-driven development process. Such an approach is invaluable in educational contexts, where the alignment of content with developmental goals and learner engagement is paramount.

The incorporation of variables related to age, gender, and thematic elements, as seen in the later versions, points to a customizable framework. This adaptability is crucial for developing content that resonates with a diverse audience, ensuring that the stories and illustrations not only entertain but also educate and foster imagination.

Lastly, the emphasis on language comprehensibility and the inclusion of friendship themes in the narratives reflect an understanding of the developmental needs of the target demographic. These elements are vital for fostering empathy and social understanding among young readers, illustrating the potential of AI-generated content to contribute positively to child development.

In conclusion, the evolution of prompts in this study reveals a thoughtful consideration of the intricacies involved in creating content for children. This progression towards more detailed, structured, and customizable prompts not only enhances the quality and relevance of the generated fairy tales and illustrations but also showcases the potential of AI as a tool for educational and developmental support. As we continue to explore this frontier, the insights gained from such iterative improvements will undoubtedly contribute to more effective and engaging storytelling strategies, tailored to the needs and imaginations of young learners.

\subsection{Evaluation - Feedback}
The effectiveness of "FairyLandAI" in generating personalized stories and accompanying images is continually assessed through both quantitative and qualitative feedback mechanisms. User engagement metrics, narrative coherence scores, and visual appeal ratings are among the quantitative measures used to evaluate the model's performance. Qualitatively, feedback from children, parents, educators, and storytelling experts is invaluable in identifying areas for improvement, refining narrative generation algorithms, and enhancing image consistency. This iterative process of evaluation and feedback ensures that "FairyLandAI" evolves in line with user expectations and pedagogical goals, maintaining its relevance and effectiveness as an educational tool.

\subsection{Limitations}
While "FairyLandAI" demonstrates significant capabilities in generating personalized fairytales for children and creating consistent character imagery for Dalle-3, certain limitations have been observed. A notable challenge arises in the model's occasional inclination to generate image prompts that may not fully align with the narrative context. In this instance, the model produced:

\noindent \textbf{Highlighted Inconsistencies:}
\begin{itemize}[label={--}]
    \item The character description in the narrative emphasized "a compassionate and wise old man," yet the image prompt generated depicted a young, adventurous character, leading to a mismatch in the narrative and visual portrayal.
    \item The narrative setting described a historical context, but the generated image prompt suggested a modern urban environment, creating a discrepancy in the story's temporal and spatial consistency.
\end{itemize}

While the first inconsistency highlights a deviation in character portrayal, the second illustrates a misalignment in setting depiction. These instances underscore the model's occasional challenges in maintaining narrative and visual consistency.

Overall, "FairyLandAI" tends to produce highly engaging and contextually appropriate narratives and images. However, ensuring consistent alignment between the narrative content and the generated image prompts remains a nuanced challenge. The optimal use of the model involves active participation from the user, especially in contexts where specific character traits and settings are crucial for the story's integrity. Continuous improvements in the model's prompt engineering and feedback mechanisms are essential in addressing these limitations.

\section{Results and Discussion}
The deployment of "FairyLandAI" in generating personalized tales for children has yielded promising results. The model's ability to craft stories that cater to the individual preferences, cultural backgrounds, and moral values of its audience signifies a notable advancement in the field of AI-driven storytelling. Through empirical evaluations involving user feedback from children, parents, and educators, several key outcomes have emerged:

\begin{enumerate}
    \item \textbf{High Engagement Levels:} Stories generated by "FairyLandAI" have shown to significantly engage young readers, as evidenced by increased reading time and positive feedback from children and parents.
    \item \textbf{Cultural and Moral Alignment:} The model successfully incorporates diverse cultural and moral values into its stories, demonstrating its sensitivity to inclusivity and ethical storytelling.
    \item \textbf{Creative and Educational Value:} Educators have highlighted the tool's effectiveness not only in entertaining children but also in teaching important moral lessons and fostering creativity.
\end{enumerate}

\section{Results and Discussion}

\subsection{Mobile Application Interface}

\begin{figure}[h!]
\centering

\begin{subfigure}{.5\linewidth}
  \centering
  \includegraphics[width=\linewidth]{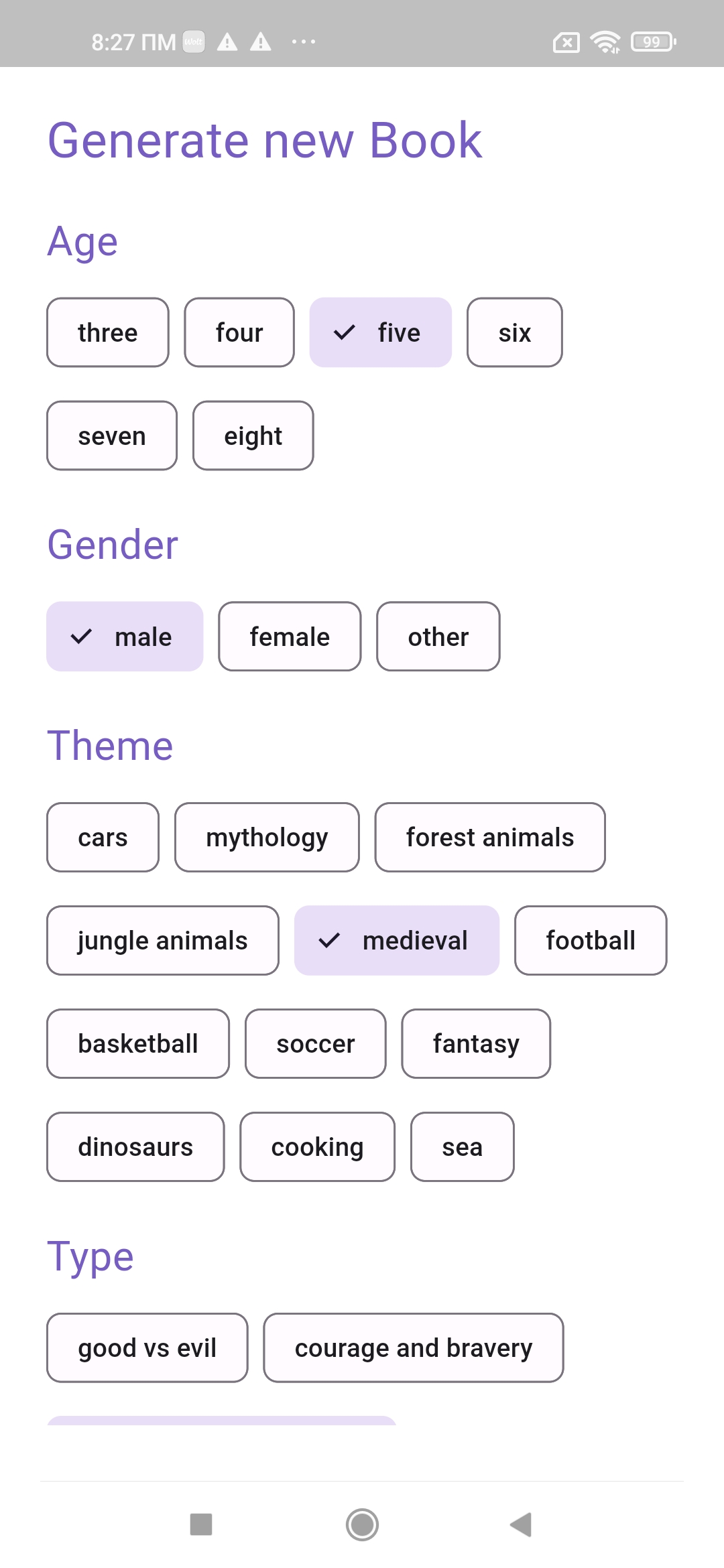}
  \caption{The Style and Language Selection screen allows users to choose the visual style and language of the generated book.}
  \label{fig:style_language}
\end{subfigure}%
% Add a % here to ensure that the figures are side by side in the same line
\begin{subfigure}{.5\linewidth}
  \centering
  \includegraphics[width=\linewidth]{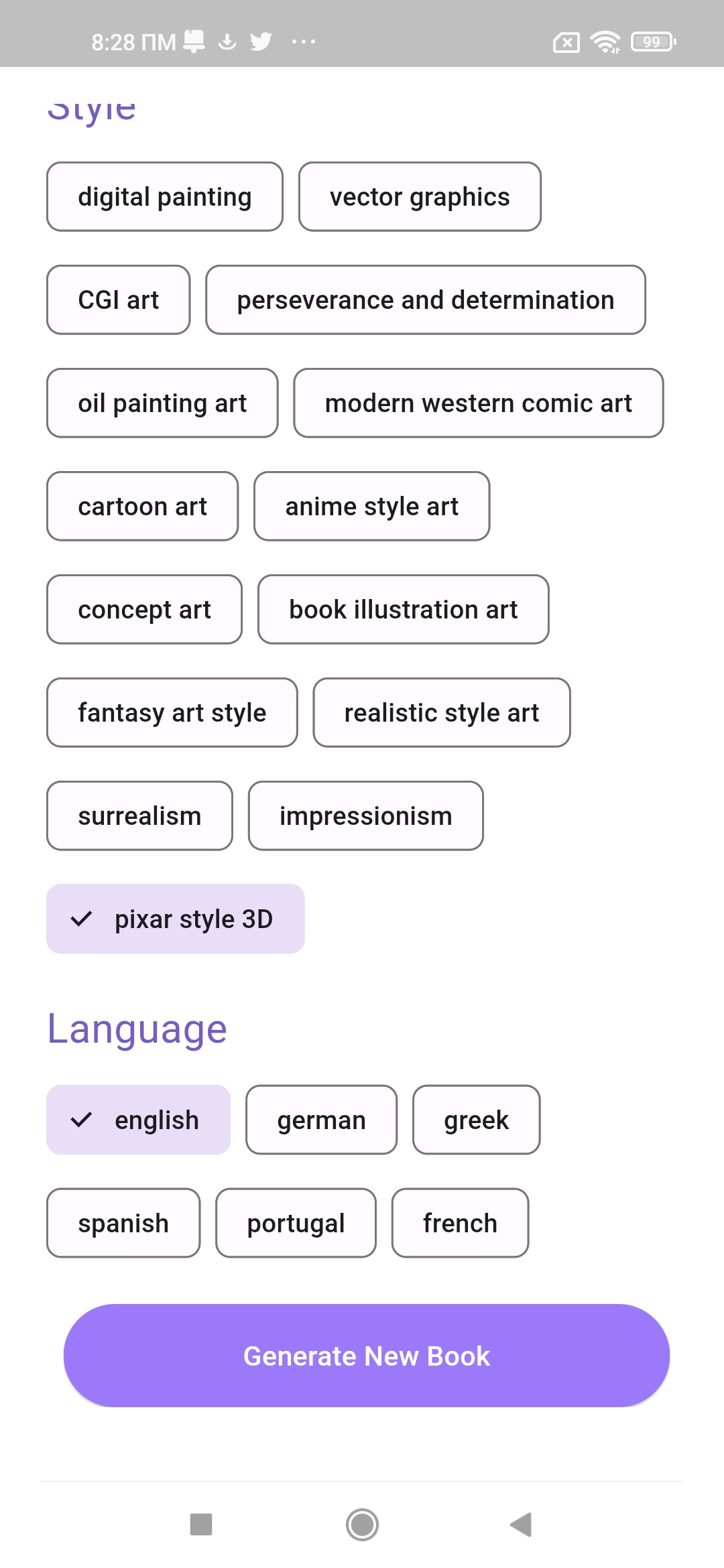}
  \caption{The Age, Gender, and Theme Selection screen presents options such as age ranges and themes like 'medieval'.}
  \label{fig:age_gender_theme}
\end{subfigure}

\begin{subfigure}{.5\linewidth}
  \centering
  \includegraphics[width=\linewidth]{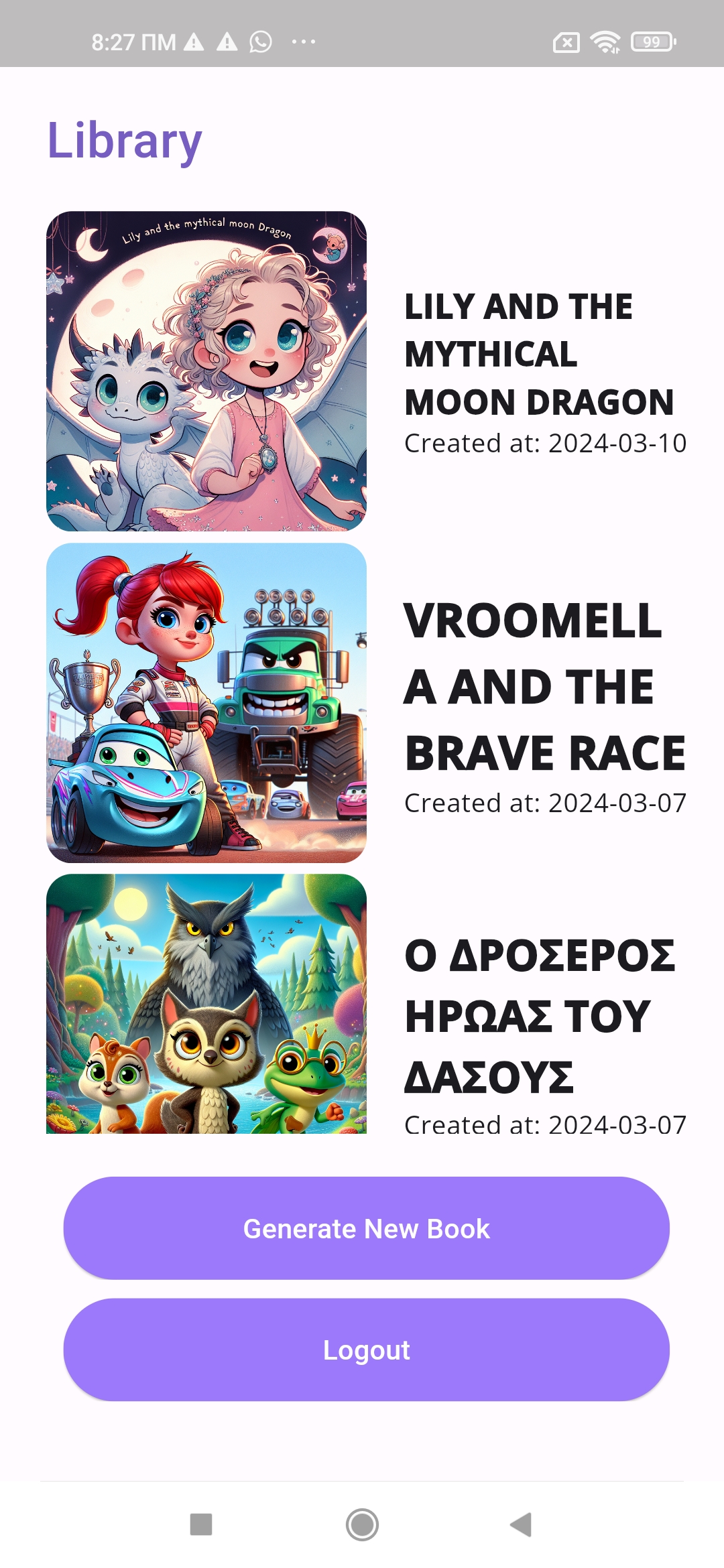}
  \caption{The Library screen displays previously created stories.}
  \label{fig:library}
\end{subfigure}%
\begin{subfigure}{.5\linewidth}
  \centering
  \includegraphics[width=\linewidth]{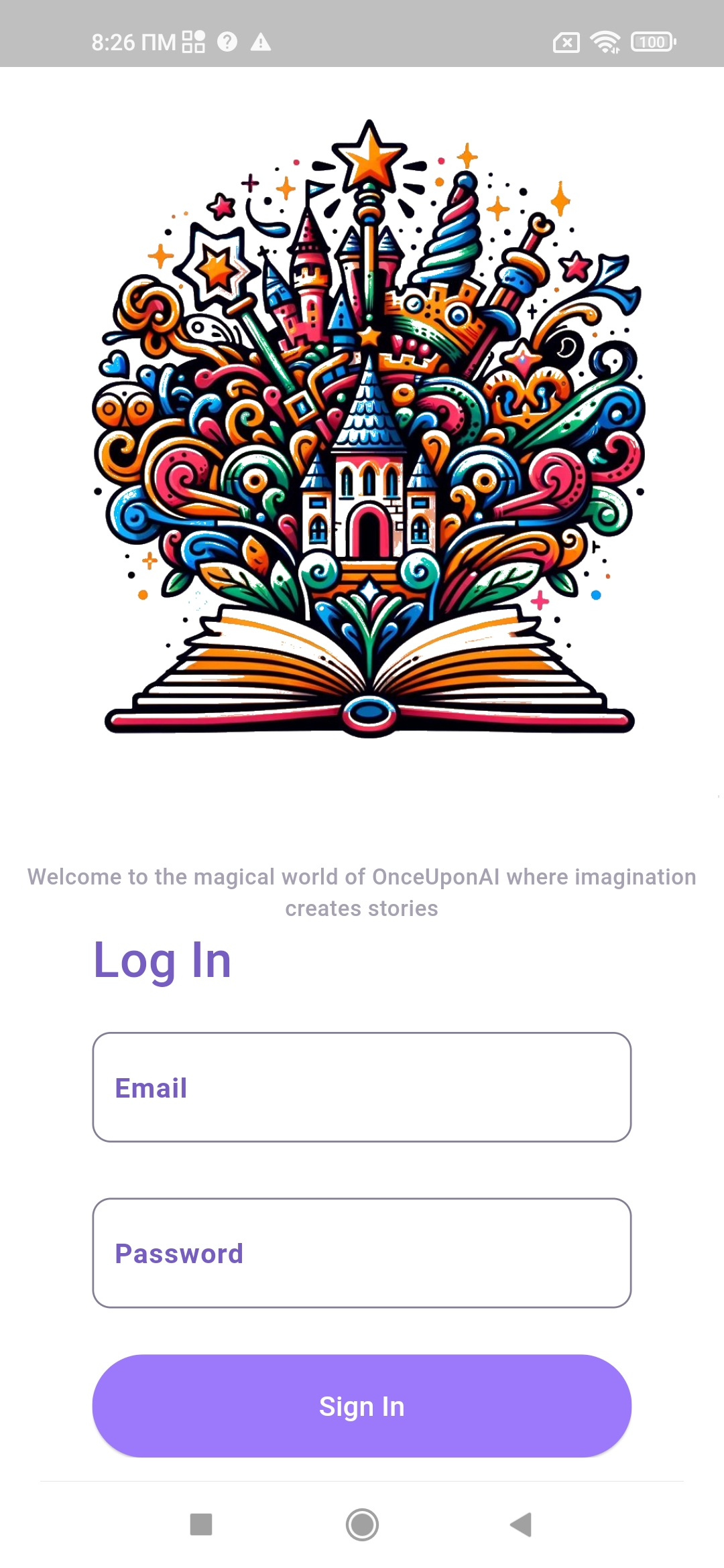}
  \caption{The Login screen provides a secure entry point to the FairyLandAI application.}
  \label{fig:login}
\end{subfigure}

\caption{Various interfaces of the FairyLandAI mobile application: (a) Style and Language Selection, (b) Age, Gender, and Theme Selection, (c) Library of created books, and (d) Secure Login screen.}
\label{fig:combined_figures}
\end{figure}

The development and design of the mobile application were driven by the need to provide an intuitive and child-friendly user experience. The following screenshots, as depicted in Figures \ref{fig:style_language}, \ref{fig:age_gender_theme}, \ref{fig:library}, and \ref{fig:login}, showcase the application's interface and functionality.

The user interface was meticulously crafted to encourage exploration and creativity. The selection screens for style and language (Figure \ref{fig:style_language}), as well as age, gender, and theme (Figure \ref{fig:age_gender_theme}), are implemented with easy-to-recognize icons and a vibrant color scheme to appeal to the app's younger audience. The library section (Figure \ref{fig:library}) showcases the interactive book covers, inviting users to revisit their favorite stories. The login page (Figure \ref{fig:login}) emphasizes the magical aspect of storytelling with its enchanting castle design, setting the stage for a creative journey.

The design choices reflect a deeper understanding of the application's target demographic. The colorful, button-based interface enables young users to make selections with ease, minimizing dependency on text-based navigation, which is crucial for users who are still developing their reading skills. Moreover, the application's capacity to customize experiences based on age and preferences indicates a tailored approach to educational technology, allowing for stories that grow with the user.

\subsection{User Engagement and Learning Outcomes}

Initial feedback from user interaction with the application has been overwhelmingly positive. Children and parents noted the ease of use, the engaging interface, and the quality of the customized stories. The feature allowing children to select themes and styles played a significant role in enhancing the user's engagement with the app. Educators praised the application's ability to incorporate language suitable for different age groups, which is critical for reading comprehension and vocabulary development.

The discussion around these results centers on the application's potential impact on early literacy development. The personalized storytelling experience not only motivates young readers but also supports language acquisition. Future studies will focus on the long-term educational benefits of using the application, assessing improvements in reading fluency and comprehension over time.

\section{Conclusion}

The FairyLandAI application exemplifies the innovative intersection of technology and education. By combining intuitive design with advanced AI-generated content, the application offers a unique platform for interactive and personalized learning. These preliminary results highlight the app's success in engaging children and the promise it holds for contributing to educational outcomes in a digital age.

Despite these positive outcomes, the challenge of aligning narrative content with generated imagery, as previously discussed, underscores the need for ongoing refinement of the model's algorithms and user input mechanisms.

"FairyLandAI" represents a significant leap forward in the use of AI for personalized storytelling, demonstrating the potential to enrich the educational and moral development of children through tailored narrative experiences. While the model showcases impressive capabilities in generating engaging, culturally sensitive, and morally aligned stories, it also faces challenges in ensuring consistency between narrative elements and generated images. The future development of "FairyLandAI" will focus on addressing these limitations through enhanced prompt engineering, deeper integration of user feedback, and continuous algorithmic improvements. By advancing the capabilities of "FairyLandAI", we aim to further bridge the gap between technology and traditional storytelling, offering a more immersive, personalized, and meaningful literary experience to children around the globe. As we move forward, the potential of AI in educational storytelling remains vast, with "FairyLandAI" leading the way in exploring new horizons for digital education and entertainment.

\section*{Acknowledgements}
% The research leading to the results presented in this paper has received funding from the Europeans Union’s funded Project HumAIne under grant agreement no 101120218.

% \appendix

% \section{Appendix title 1}
% %% \label{}

% \section{Appendix title 2}
% %% \label{}

\bibliographystyle{elsarticle-harv} 
\bibliography{example}



%% For example:

%% \bibitem[Aladro et al.(2015)]{Aladro15} Aladro, R., Martín, S., Riquelme, D., et al. 2015, \aas, 579, A101

%%\end{thebibliography}

\end{document}